\documentclass{article}
\usepackage{times}
\usepackage{soul}
\usepackage{url}
\usepackage[hidelinks]{hyperref}
\usepackage[utf8]{inputenc}
\usepackage[small]{caption}
\usepackage{graphicx}
\usepackage{amsmath}
\usepackage{booktabs}
\usepackage{algorithm}
\usepackage{algorithmic}
 \usepackage{natbib}
\usepackage{amssymb}
\usepackage{subfig}

\title{Deep Learning for Multi-Scale Changepoint Detection in Multivariate Time Series}

\author{Zahra Ebrahimzadeh \and Min Zheng \and Selcuk Karakas \and Samantha Kleinberg
\\ Stevens Institute of Technology}

\begin{document}

\maketitle

\begin{abstract}
Many real-world time series, such as in health, have changepoints where the system's structure or parameters change. Since changepoints can indicate critical events such as onset of illness, it is highly important to detect them. However, existing methods for changepoint detection (CPD) often require user-specified models and cannot recognize changes that occur gradually or at multiple time-scales. To address both, we show how CPD can be treated as a supervised learning problem, and propose a new deep neural network architecture to efficiently identify both abrupt and gradual changes at multiple timescales from multivariate data. Our proposed pyramid recurrent neural network (PRN) provides scale-invariance using wavelets and pyramid analysis techniques from multi-scale signal processing. Through experiments on synthetic and real-world datasets, we show that PRN can detect abrupt and gradual changes with higher accuracy than the state of the art and can extrapolate to detect changepoints at novel scales not seen in training.
\end{abstract}

\section{Introduction}
Changepoints, when the structure or parameters of a system change, are critical to detect in many domains. In medicine, finance, climate science and other fields, these changes can indicate that important events have occurred (e.g. onset of illness or a financial crisis), or that a system has changed in critical ways (e.g. increasing illness severity). Both types of changes will influence decisions about treatment and policies. Changepoint detection (CPD) aims to find these critical times automatically, but this is a challenging problem as changes can result in complex patterns across multiple observed variables, which may also be interdependent. Further, not all changepoints lead to a sudden transition; many occur over a period of time (e.g. weightloss, transition between activities) and are harder to identify.  Being able to detect such changes, though, will have wide applicability in many domains.
 
Both parametric and nonparametric solutions have been proposed for CPD. Parametric methods \citep{adams2007bayesian,montanez2015inertial} often make strong assumptions about the data and most are context specific, so they face difficulty when changes result in complex temporal patterns that are hard to model manually. 
Nonparametric methods \citep{saatcci2010gaussian,li2015m} address this with engineered divergence metrics or kernel functions, but the choice of parameters or kernels significantly affects accuracy. More fundamentally, these methods focus on abrupt changes, while in real world applications like activity recognition or finding onset of illness, a change may be gradual and happen over different durations. 
Some methods exist to detect gradual changepoints \citep{bardwell2017bayesian,harel2014concept}, but cannot handle changes occurring at arbitrary timescales. In applications such as detecting activity changes, though, how quickly someone transitions from sitting to standing should not affect CPD accuracy.

At the same time, Deep Neural Networks (DNN) can learn functions automatically and be adapted to new tasks if there is sufficient training data, leading to use in applications such as time series classification \citep{yang2015deep}. DNNs have not yet been exploited for CPD, though, and face challenges in generalizing across scales (requiring training data for all possible transition speeds). Since such data can be costly or infeasible to collect, it is ideal to have a scale-invariant approach that can generalize beyond observed timescales.

To address the gap in CPD, we propose a novel DNN architecture for CPD in multivariate data. Our approach further makes two key contributions to neural network architecture (trainable wavelet layer, Pyramid recurrent neural networks (PRN)), which provide scale invariance and allow better use of multivariate data where patterns appear at varying speeds. We focus here on CPD due to the significance of this task, however the approach is highly general and may be applicable to classification problems and time series analysis more generally. On both simulated and real-world data, we show that our architecture allows more accurate detection of both abrupt and extremely gradual (e.g. like weight loss) changes, and further is scale invariant -- allowing detection of changes at any timescale, regardless of those seen in training.

\section{Related Work}

\textbf{Changepoint detection} is a core problem for time-series analysis. One approach is to use a model and find  when observations deviate from what is predicted by the model. Bayesian Online CPD (BOCPD) \citep{adams2007bayesian} detects changes in an online manner, but makes the limiting assumption that the time series between changes has a stationary exponential-family distribution. More generally, Bayesian techniques require full definition of the likelihood function \citep{montanez2015inertial}, which may be difficult to specify. Nonparametric models \citep{saatcci2010gaussian} increase flexibility, but also increase computational complexity. Gaussian Graphical Models (GGMs) move beyond the univariate case to detect changes in multivariate time series \citep{xuan2007modeling}. GGM is an offline method that models the correlations between multivariate time series using a multivariate Gaussian. This method is closest to ours as it focuses on multivariate CPD, but unlike our approach it makes strong assumptions about the data distribution.

To eliminate the need to specify a model, model-free approaches have emerged, such as density-ratio estimation methods \citep{yamada2013change,liu2013change,kuncheva2014pca} and kernel methods \citep{harchaoui2009kernel,li2015m}, but these depend strongly on the chosen estimation methods or kernels, which might be domain specific. \citep{ide2016change} proposed an online method for CPD in multivariate data, which focuses on handling noisy data by first extracting features to capture major patterns. While this can overcome noise, it may miss more subtle or gradual changes. 
Other techniques define custom divergence functions such as using the difference in a covariance matrix \citep{barnett2016change}, however not all changes will result in a significant change in covariance, such as when there are small changes across a number of variables that in aggregate indicate a system change. Statistical methods have other limitations such as reliance on the choice of kernels in MMD \citep{gretton2007kernel}, choice of parameters in Hotelling T-square \citep{chen2000parametric}, or prior information in CUSUM \citep{page1954continuous}; or high computational complexity with large samples for generalized likelihood ratio  \citep{james1992asymptotic}. Thus such models cannot be readily used in a new domain without re-engineering the divergence or kernel functions. 

Few methods were explicitly designed to detect gradual changes, though BOCPD has been extended this way by reformulating changes as segments instead of points \citep{bardwell2017bayesian}. Alternatively, gradual changes can be formulated as concept drifts \citep{harel2014concept}. We instead develop scale invariant models, which  can handle short- and long-term temporal patterns, and can generalize to novel time-scales without extra computation or training data. 

\textbf{Deep learning}, which allows recognition of complex patterns in large datasets without engineering of features and metrics, provides a promising way to address the core challenges of CPD. CNNs can learn to extract increasingly abstract features from raw data through a stack of non-linear convolutions, allowing, for example, recognition of  hundreds of object types in natural images \citep{szegedy2015going}. RNNs further can learn complex temporal patterns in sequences of arbitrary length (e.g. to recognize human activities \citep{hammerla2016deep}), which are exactly the types of changes that are challenging for CPD.

Ideally, a CPD method should perform equally well on test data regardless of whether changes happen faster or slower than seen in training data. However, the fixed resolution of CNN and RNN architectures makes them sensitive to scale. While CNN extensions can model multiple scales simultaneously \citep{shen2015multi}, this is not the same as scale invariance, as it simply concatenates features. \citet{chung2016hierarchical} introduce Hierarchical Multiscale Recurrent Neural Networks (HM-RNN), which  process a sequence through successive RNN layers at different resolutions. This can improve efficiency by detecting boundaries and only updating the RNN when a change is recognized. However, the layers of RNN resemble layers of convolution in CNNs (modeling the signal at a different abstraction level) and are not invariant to scale changes at the same abstraction level. Our proposed architecture, PRN, in contrast combines advantages of both CNN and RNN and augments them with scale invariance.

RNNs are naturally built to model long-term dependencies, as is necessary for recognizing gradual changes, but suffer from vanishing gradients. Extensions such as Long Short-Term Memory (LSTM) \citep{hochreiter1997long}  solve vanishing gradients, but still have limited memory space. Intuitively, information from an infinitely long sequence cannot be stored in a fixed-dimensional RNN cell. Skip RNNs were proposed to reduce computational complexity by skipping state updates while preserving the performance of baseline RNN models \citep{campos2017skip}. The skip operation avoids redundant updates, but risks skipping temporal dependencies (especially long term ones), which can hurt the overall performance. 
To address this, recent work has augmented RNNs with various types of memory or stack \citep{sukhbaatar2015end,joulin2015inferring}, but these methods are not scale-invariant. 
Frameworks like Feature pyramid networks \citep{lin2017feature} and wavelet CNN \citep{fujieda2018wavelet} have  been proposed for images with different scales or resolutions, though neither are directly applicable to time series data. In contrast, our PRN models infinitely long sequences with its multi-scale RNN, which forms a stack of memory cells in an arbitrary number of levels. A higher-level RNN cell in a stack has lower resolution, and thus can store longer dependencies at no additional computational cost, while a lower-level RNN cell has a high resolution and prevents the loss of details in the short term. 

\section{Method}
We propose a new class of deep learning architectures called Pyramid Recurrent Neural Networks (PRNs). 
The model takes a multi-variate time series and transforms it into a pyramid of multi-scale feature maps using a trainable wavelet layer (NWL). All pyramid levels  are processed in parallel using multiple streams of CNN with shared weights, yielding a pyramid of more abstract feature maps (DWN). Next, we build a multi-scale RNN on top of the pyramid feature map, to encode longer-term, dependencies. The PRN output is used to detect changes at each time step with a binary classifier.

\subsection{Deep Wavelet Neural Networks (DWN)}

CNNs can learn to recognize complex patterns in multivariate time series, partly due to parameter-sharing across time, which leads to shift-invariance. CNNs are not scale-invariant, though, so a learned pattern cannot necessarily be recognized when it appears more gradually or quickly. To make CNNs scale invariant, we introduce Deep Wavelet Neural Networks (DWN), which consist of a Neural Wavelet Layer (NWL) followed by parallel streams of CNN.

\begin{figure}[t]
\centering 
\includegraphics[width=\linewidth]{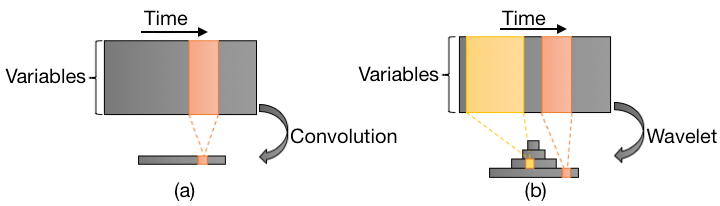}
\caption{\label{fig:wavelet} Illustration of (a) a convolutional layer; (b) the Neural Wavelet Layer. One feature map of each output is shown.}
\end{figure}

The NWL can be seen as a set of multi-scale convolutions with trainable kernels, which are applied in parallel on each variable of the input time series. The input to the NWL is a multivariate time series, $X \in \mathbb{R}^{T \times c}$, where $T$ is the number of timepoints and $c$ is the number of variables. The NWL takes $X$ and produces multiple feature maps, which together form a pyramid of convolution responses. That is:
\begin{equation}
f_{NWL}(X) = (H_1, H_2, ... , H_k): H_i \in \mathbb{R}^{T/2^{i-1} \times c}.
\end{equation}
An example is shown in Figure \ref{fig:wavelet}. Specifically, the NWL uses the filter bank technique \citep{mallat1999wavelet} for discrete wavelet transform. Given a pair of separating convolutional kernels (typically a low-pass and a high-pass kernel), it convolves the signal with both, outputs the high-pass response, and down-samples the low-pass response for the next iteration. It repeats this process and in each iteration outputs an upper level of the output pyramid. Although traditional wavelets such as Haar or Gabor \citep{mallat1999wavelet} can be used, we have experimentally found that initializing the filter banks with random numbers and training them using backpropagation with the rest of the network leads to higher accuracy.

More formally, the NWL is characterized by its trainable kernels $K_l^{(v)},K_h^{(v)}\in \mathbb{R}^{\tau \times c}$ for all variables $v\in\{1 ... c\}$, where $\tau$ is the kernel size. Given each channel of $X$ as input (e.g. $X^{(v)}$), the NWL iteratively computes lowpass and highpass responses, starting with $L_1^{(v)}$ and $H_1^{(v)}$, that are:
\begin{equation}
L_1^{(v)} = \omega(X^{(v)} * K_l^{(v)}) \quad , \quad H_1^{(v)} = \omega(X^{(v)} * K_h^{(v)}), 
\end{equation}
where $*$ is convolution and $\omega$ is a downsampling operation (e.g. implemented by linear interpolation). At the $i$-th iteration of the wavelet transform, given $L_{i-1}^{(v)}$ and $H_{i-1}^{(v)}$, it computes $L_{i}^{(v)}$ and $H_{i}^{(v)}$ such that:
\begin{equation}
L_{i}^{(v)} = \omega(L_{i-1}^{(v)} * K_l^{(v)}) \quad , \quad U_{i}^{(v)} = \omega(L_{i-1}^{(v)} * K_h^{(v)}) 
.
\end{equation}
This operation is repeated for a pre-specified number of times, $k$, or until the length of $L_i^{(v)}$ and $H_i^{(v)}$ becomes smaller than a threshold. The hyperparameter, $k$, can be selected using cross-validation. A larger $k$ (or smaller threshold) results in a larger receptive field at the highest level of the pyramid, enabling the detection of more gradual patterns. However, a large $k$ also brings more computation and requires a larger buffer in the case of online processing. 

The output of each iteration $i\in\{1 ... k\}$ for variables $v\in\{1 ... c\}$ can be concatenated to form 
\begin{equation}
L_i=[L_i^{(1)}|L_i^{(2)}|...|L_i^{(c)}]
, \quad
H_i=[H_i^{(1)}|H_i^{(2)}|...|H_i^{(c)}],
\end{equation}
where $[.|.]$ indicates concatenation. The output of the NWL is the stack of all $H_i$. These are called different \textit{levels} of a \textit{pyramid} throughout this paper. In the original filter bank method the last lowpass response, $L_k$, is also stacked with the output but we did not observe an improvement with $L_k$.  

The key advantage of a NWL over a conventional convolution layer is that a single wavelet can encode the input with multiple granularities at once, whereas a single convolution only encodes a single granularity. Although different layers of a CNN have different granularities, they encode the data at a different level of abstraction, and thus cannot simultaneously extract the same pattern at different scales. On the other hand, a single wavelet layer can encode changes with the same patterns at different paces, simultaneously into the same feature map, at different levels of the pyramid.

We will use the proposed NWL as a part of a larger, deeper architecture, which is described in the rest of this section. Hence, an important aspect of NWL is that it can be used as a layer of a deep network, in composition with other neural layer types such as convolutional and fully connected layers. For example, the input to a wavelet layer can be the output of a convolutional layer. Alternatively, to stack a convolutional layer on the output of a wavelet layer, one should apply the convolution on each level of the wavelet pyramid, resulting in a pyramid-shaped output. 

Accordingly, a network composed of one wavelet layer and an arbitrary number of other layers, can take a multivariate time series as input, and produce a pyramid-shaped response as output. We refer to such a network architecture as a Deep Wavelet Neural Network (DWN). In this paper we use a specific form of DWN, which starts with a NWL, directly applied on the input time series $X$, followed by parallel streams of CNN with shared parameters, each of which takes one level of the NWL pyramid. More specifically, we use an $\ell$-layer CNN with a down-sampling stride of $p_j$ at the $j$-th layer, which results in a total down-sampling factor of $P=\prod_{j=1}^\ell p_j$, and with $f_j$ feature maps at the $j$-th layer. We apply that CNN in parallel on each level of the output pyramid of the NWL, which means for each $i\in\{1 ... k\}$, it gets $H_i \in \mathbb{R}^{T/2^{i-1} \times c}$ and outputs $C_i \in \mathbb{R}^{T/2^{i-1}/P \times f_\ell}$.

\begin{figure}
\centering 
\includegraphics[width=\linewidth]{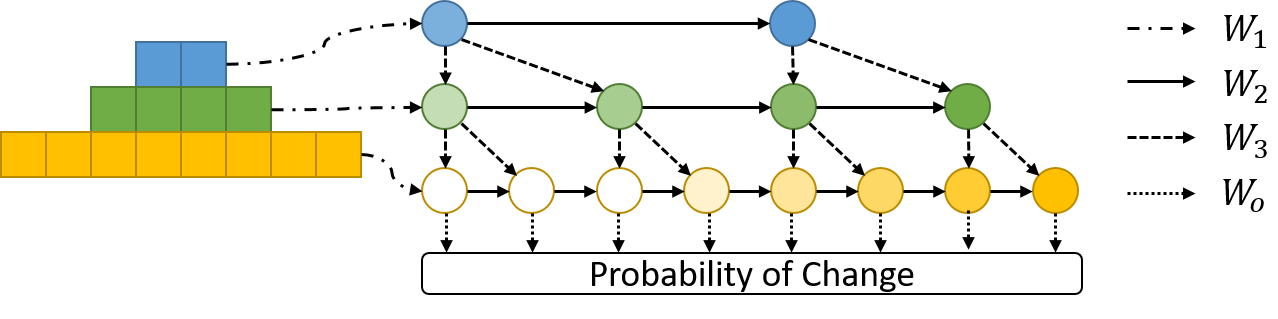}
\caption{Pyramid Recurrent Layer with downsampling ratio of 2.}
\label{fig:rnn}
\end{figure}

\subsection{Pyramid Recurrent Layer}

The DWN output is a multi-scale pyramid of sequential feature maps that encode short-term temporal patterns at different times and scales. It is common to process sequential features using an RNN, to encode longer-term temporal patterns. However, conventional RNNs process a single sequence, not a multi-scale pyramid of sequences. Similar to the need for a wavelet layer, RNNs are not scale-invariant, so an RNN cannot necessarily recognize a temporally shortened or stretched instance of a learned pattern without having seen this scale in  training. Further, RNNs fail to learn very gradual patterns, due to limited memory. While this can be addressed by memory-augmented networks, they remain sensitive to scale. 

To address these issues, we introduce a novel hierarchically connected variant of RNNs. Our proposed network, PRN, scans the multi-scale output of a DWN, and simultaneously encodes temporal patterns at different scales. 
An RNN is applied in parallel on different levels of the input pyramid. On each level at each step, it takes as input the corresponding entry from the input pyramid, along with the most recent output of the RNN operating at the upper level. We concatenate those two vectors and feed as input to the RNN. We refer to this technique as Pyramid Recurrent Layer (PRL).

Denoting the value at level $i$ of the input pyramid at time $t$ as $C_i[t]$, and assuming the downsampling ratio in the wavelet transform is $d$, (i.e., each level of the pyramid has $d$-times the length of its upper level) we can write the recurrent state at level $i$ and time $t$ as:
\begin{equation}
h_i[t] = \sigma(W_1 C_i[t] + W_2 h_i[t-1] + W_3 h_{i+1}[\left \lfloor{t/d}\right \rfloor ]+b)
,
\end{equation}
where $\sigma$ is a nonlinear activation function such as ReLU, and $W_1$, $W_2$, $W_3$ and $b$ are trainable parameters of this layer. These parameters define a linear transformation of the current state, past state, and higher-level state, as illustrated in Figure \ref{fig:rnn}. Note that the proposed hierarchical structure is agnostic of the function of each cell. Although we used a simple RNN cell for illustration, we could use any variant of RNNs such as LSTM \citep{hochreiter1997long} or Skip RNN \citep{campos2017skip} as the RNN cell. 

The proposed architecture can be compared with an RNN operating on a single data sequence. If the data granularity is high, the RNN likely fails to model long-term dependencies, due to the well-known problem of vanishing gradients. One can lower the data granularity, so long-term patterns can be summarized in fewer steps, but this results in the loss of details. Accordingly, conventional RNNs were not designed to effectively detect both abrupt and gradual patterns at the same time. 
On the other hand, in the proposed PRL, each RNN unit is provided with inputs from the same level of granularity as well as the level above. The RNN that operates at the lowest level, in turn, receives information from all levels of granularity. Figure \ref{fig:rnn} illustrates the effect of forgetting using decreasing color saturation. While it is impossible to keep track of the past through the lower level alone, the information path from upper levels connect the past to present in only three steps. This lets the PRL model long-term patterns, while it can still model fine details through the lower levels.

\subsection{Pyramid Recurrent Neural Networks (PRN)}

We propose PRN as a composition of a DWN and a PRL. An input time series of arbitrary length is transformed through a DWN into a pyramid-shaped representation, which is then fed into a PRL. For CPD and other classification problems, a logistic regression layer is built on the output of the RNN cells that operate at the lowest level of the pyramid. This layer produces detection scores at each time step with the highest possible granularity. 
The detection score for time $t$ is:
\begin{equation}
y_t = \sigma(W_o h_1[t] + b_o) 
,
\end{equation}
where $\sigma$ is the sigmoid function and $W_o$ and $b_o$ are trainable parameters. The classification loss at each time is the cross entropy loss, where $y^*_t$ is the ground truth at time $t$:
\begin{equation}
E_t = y^*_t \log{y_t} + (1-y^*_t) \log{(1-y_t)}
,
\end{equation}
We optimize this loss using stochastic gradient descent on parameters of the classifier ($W_o$ and $b_o$), PRL ($W_1$, $W_2$, $W_3$ and $b$), and NWL ($K_l$ and $K_h$).

\section{Evaluation}
We compare the proposed PRN to deep learning and CPD baselines. Using both simulated and real-world datasets, we show that PRNs can detect abrupt and gradual changes more accurately than existing methods and can be used for activity recognition by learning labels for different changes.

\subsection{Datasets\label{subsec:dataset}}

\begin{figure}[t]
\centering 
\includegraphics[width=0.75\textwidth]{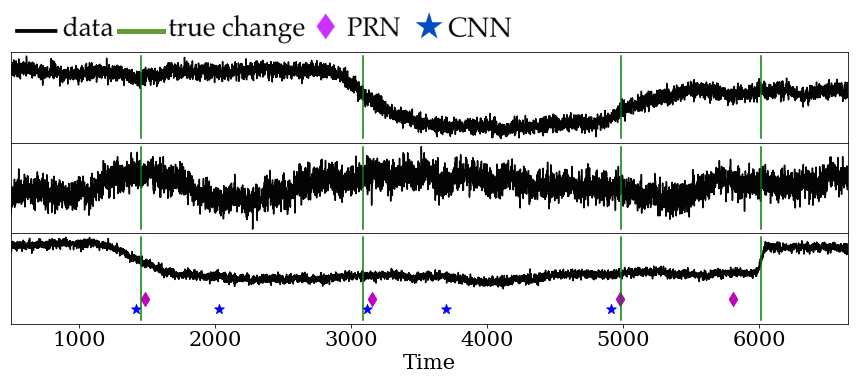} 
\caption{Results for 3 of 12 dimensions of the synthetic dataset.}
\label{fig:vis-synth}
\end{figure}

\textbf{Synthetic dataset} 
We create a synthetic dataset to evaluate accuracy at simultaneously detecting gradual and abrupt changes. We construct 2000 time series each with 12 variables and 8192 time steps. 
Each time series is a combination of a Brownian process and white noise. Each has 4 randomly placed changepoints, which are defined as a shift in the mean of 4 randomly selected dimensions with random duration and size of change. Duration 0 is an abrupt change, while longer ones provide more challenging cases to recognize. An example of the simulated time series together with ground truth and detection results are shown in Figure \ref{fig:vis-synth}. We randomly split the series, with 1000 for training and 1000 for testing. To demonstrate robustness of the proposed method against variability in scale, we also do a split by scale, where all changes in one half are strictly more gradual than all in the other half.

\textbf{OPPORTUNITY dataset} This activity recognition dataset \citep{chavarriaga2013opportunity} provides a challenging real-world test, as activity changes take place at varying durations. The data are on-body sensor recordings from 4 participants performing activities of daily living, such as cleaning a table. Each participant has 6 records (runs) of $\sim$20min each. Values of 72 sensors (10 modalities) were recorded at 30Hz, and manually labeled with 18 activities. Following \citep{hammerla2016deep}, we ignore variables with missing values, which leads to 79 variables for each record. We use run 2 of subject 1 for validation and runs 4 and 5 of subjects 2 and 3 for test, and the rest for training. We consider the transition between two activities a changepoint, and use the activity labels from the data as ground truth. 

\textbf{Bee Waggle Dance dataset} is our second real-world test case \citep{oh2008learning}. 
The data includes six videos of bee waggle dances (used to communicate with other bees) with 30 frames per second. The data include 3 variables encoding the bee's position and head angle at each frame. Using the position and angle information, each frame is labeled as ``turn left'', ``turn right'', or ``waggle dance.'' Similar to  OPPORTUNITY, we consider the transition between two activities a change point. We test our method and other baselines on sequence 1 of the bee data, training on time series from the first 256 frames and testing on the other 768 frames. We use this small training data for consistency with prior works \citep{saatcci2010gaussian}, and as a challenging evaluation. 

\subsection{Baselines}
We compare the proposed architecture to unsupervised CPD and supervised deep-learning baselines: 

\textbf{GGM} \citep{xuan2007modeling} is related to BOCPD, a classic method for CPD, but was selected for fairer comparison as GGM is offline and allows multivariate time series.

\textbf{CNN} We use a CNN that takes a time series as input and predicts a sequence of detection scores for changes. Due to the widely used max-pooling layers, the output has a lower temporal granularity compared to the input. We denote the ratio of output length to the input length as $\gamma$.

\textbf{RCN} We apply an RNN to the output of the CNN. The output has the same granularity as CNN, while each step of the output has a larger receptive field that encodes past data.

\textbf{HM-RNN} We compare against HM-RNN \citep{chung2016hierarchical}, which is multiscale and more efficient than RNN.

\textbf{DWN} Our DWN is formed by applying an NWL to the input time series and feeding the output pyramid levels to parallel branches of a CNN. The output of CNN is upsampled to have the same size and fused by arithmetic mean. 

\textbf{PRN} We apply the complete proposed method which consists of a DWN followed by a Pyramid Recurrent Layer to fuse levels of the pyramid.

\textbf{PRN-S} We replace the standard RNN cell in our PRN with a Skip RNN \citep{campos2017skip} to test whether a more efficient RNN can provide the same performance as PRN.

\subsection{Implementation details}
We briefly summarize implementation of the baselines, and provide full details in the appendix. All of the deep-learning baselines share a core CNN architecture, with each convolution layer followed by max-pooling and ReLU activation, and output fed to a fully connected perceptron with sigmoid activation, which results in binary detection scores at each time step. 
The granularity ratio $\gamma$ for this architecture is $1/16$. 
For DWN, PRN, and PRN-S, we used a 7-level wavelet with kernel size 3 for both synthetic and OPPORTUNITY data. Due to the small size and more abrupt transitions in the bee data, we used a 5-level wavelet with kernel size 3. For all datasets RCN and PRN used an LSTM cell with 256 hidden units, and 128 units for HM-RNN (3 layers). 
At test time, the models take a time series and predict a sequence of detection scores. To detect changepoints, we apply non-maximum suppression with a sliding window of length $\omega$ (which controls how nearby two distinct changes can be) and filter the maximum values with a threshold. We evaluate AUC by iterating over this threshold. 
For GGM, we use the full covariance model to capture the correlations between features. We use a uniform prior as in \citep{xuan2007modeling}, and set the pruning threshold to $10^{-20}$. GGM is unsupervised, but used the same test data as all other methods for fair comparison.

The real world datasets (Bee and OPPORTUNITY) include diverse changepoints formed by transitions between many activities. Thus we use multitask learning, training the model to both detect changes and classify activities by changing the output dimension of the last fully connected later to have $N$ units (N=activities+1: 19 for OPPORTUNITY, and 4 for Bee), with the first N-1 units predicting a log probability for each activity and the last unit the probability of a change.

As detected changepoints may not exactly match the true times, we use a tolerance parameter $\eta$ that sets how close a detected change must be to a true change to be considered correct. 
Precision is matched detections divided by all detections, and recall is matches divided by true changes. 

\begin{figure*}
\centering 
\subfloat[Train abrupt, test gradual]{
\includegraphics[width=0.28\textwidth]{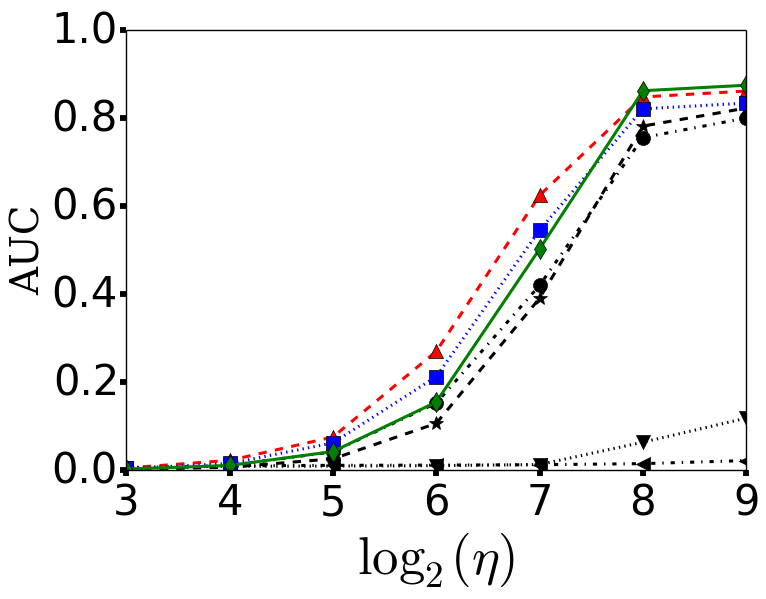} 
\label{fig:auc-synth-a2g}}
\subfloat[Train gradual, test abrupt]{
\includegraphics[width=0.28\textwidth]{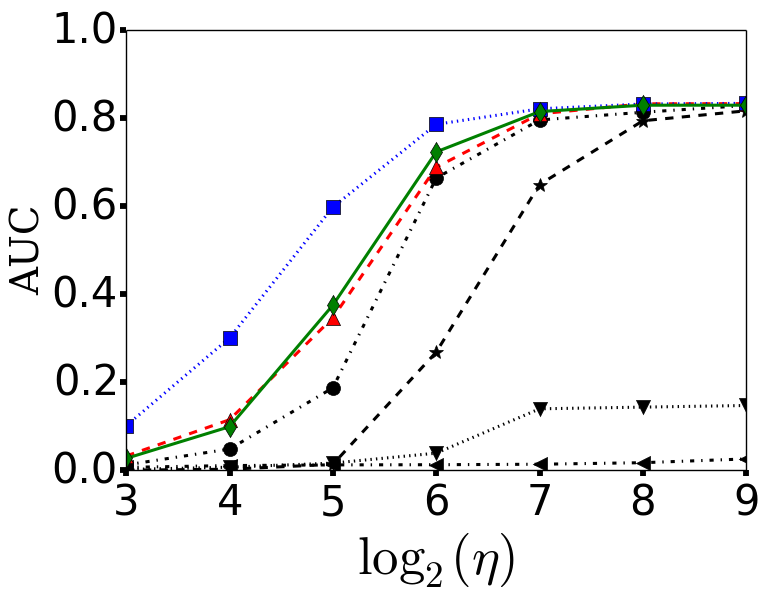} 
\label{fig:auc-synth-g2a}
}
\subfloat[Train all, test all]{
\includegraphics[width=0.28\textwidth]{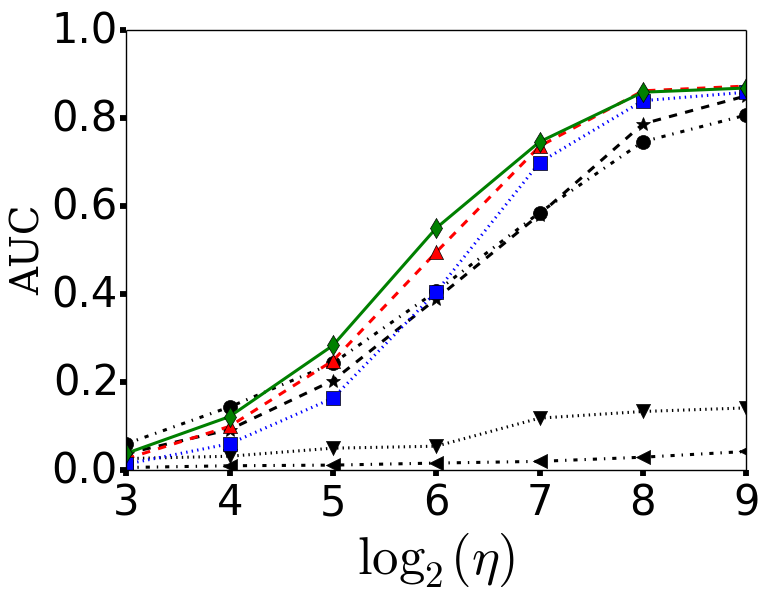} 
\label{fig:auc-synth-mix}}
\subfloat[legend]{
\includegraphics[width=0.08\textwidth]{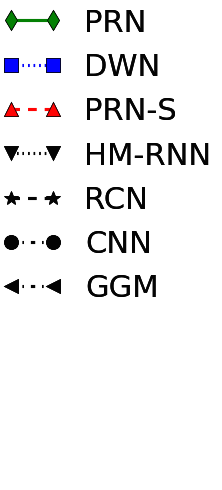} 
\label{fig:auc-syn-legend}}
\caption{\label{fig:auc-synth} AUC (Area Under the ROC Curve) results for the synthetic dataset with three different test/train scenarios. $\eta$ is the tolerance for how close in time a detected change must be to a true change to be considered a positive. See Appendix for AUC score tables.}
\end{figure*} 

\subsection{Results}

\subsubsection{Synthetic data}
Fig. \ref{fig:vis-synth} illustrates results for our scale invariant PRN and scale sensitive CNN. CNN has a higher false positive rate, while also missing a change. While detected changes and ground truth are not always precisely aligned, small gaps are acceptable for gradual changes, where it can be hard to define a single moment when the change occurs.

We use three train-test splits to test extrapolation (from abrupt to gradual and vice versa) and ability to handle a mix of scales. First, real-world cases are likely to have a mix of scales in both training and test data, and in this case (fig. \ref{fig:auc-synth-mix}) PRN and PRN-S perform best, while HM-RNN and GGM have the lowest AUC. HM-RNN relies on correct detection of boundaries and is not invariant to scale changes at the same level of abstraction, leading to errors with gradual changes. GGM highlights the challenge of unsupervised learning with a small number of events per time series. 
In the two scale-variant splits, the model must extrapolate patterns from training data to novel scales. Comparing mixed scales to training on abrupt and testing on gradual changes (fig.\ref{fig:auc-synth-a2g}), we see that this is challenging for methods that are not  scale-invariant, as shown by the drop in performance for CNN (from 41\% to 15\%) and RCN (from 39\% to 11\%) when the tolerance is 64 steps ($2^6$).  
While AUC of our methods (DWN and PRN) also decreases with this more difficult task, the drop is substantially lower for DWN (20\%) due to the wavelet layer and shared parameters across scales. (See Appendix for details). 
Finally, when training on gradual and testing on abrupt changes (fig. \ref{fig:auc-synth-g2a}) our approach again outperforms CNN and RCN  due to its ability to generalize across scales. With $\eta =64$, AUC is higher for PRN (72\%) and DWN (79\%) compared to CNN (67\%) and RCN (27\%). 
Note that DWN outperforms PRN and PRN-S, and similarly, CNN outperforms RCN, as recurrent architectures are generally less effective for this kind of extreme generalization. The performance of our DWN shows the effectiveness of the added wavelet layer in modeling both gradual and abrupt changes in time series.

\begin{figure}
\centering 
\subfloat[Opportunity data]{
\includegraphics[width=0.5\textwidth]{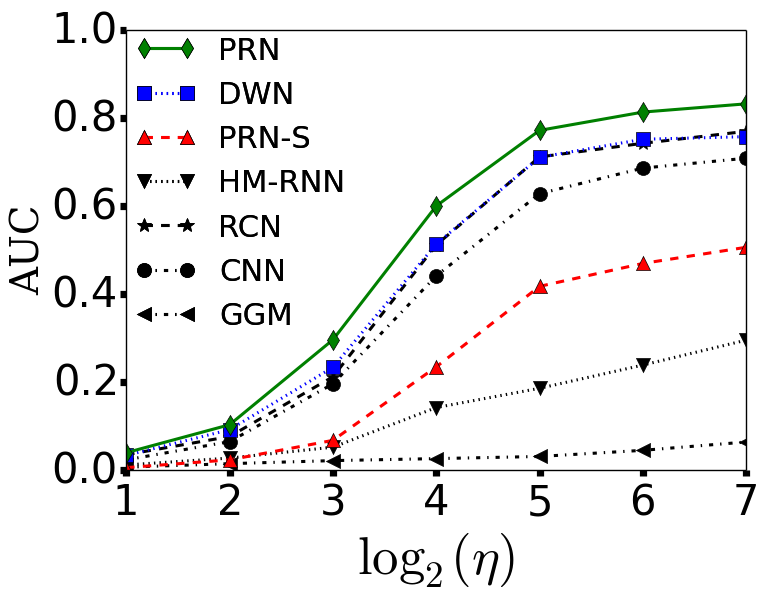} 
\label{fig:auc-opp}
}
\subfloat[Bee Waggle Dance Data]{
\includegraphics[width=0.5\textwidth]{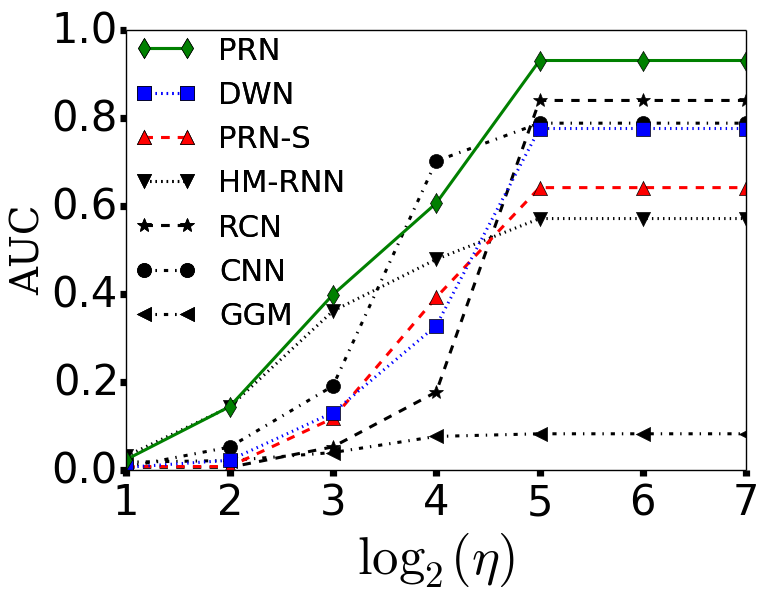} 
\label{fig:auc-bee}
}
\caption{Results for real-world data. $\eta$ has a unit of 1/30sec for both datasets. See Appendix for AUC score tables.}
\label{fig:real}
\end{figure}  

\subsubsection{Opportunity data}
On this real-world activity data, PRN outperforms all other methods at all tolerance levels, as shown in fig. \ref{fig:auc-opp}. In contrast to the synthetic data, PRN-S now has significantly lower AUC than both PRN and DWN. It may be that Skip RNN is skipping important information encoded in our wavelet later. HM-RNN has the lowest AUC of all RNN based methods, as it relies on distinct differences in distribution, which may not happen between activities. GGM had the lowest AUC for all cases, showing that supervised methods that can leverage data labels can provide better CPD. When the tolerance is 2sec ($\eta=2^6=64$), a reasonable value for activity recognition, PRN has 82\% AUC while DWN, RCN, CNN, PRN-S, HM-RNN, and GGM respectively achieve 75\%, 74\%, 69\%, 47\%, 24\%, and 5\%. Full results are in the Appendix.

The deep learning methods, PRN, PRN-S, RCN, DWN, HM-RNN, and CNN, respectively took 110, 105, 80, 69, 24, and 6 minutes to train and converge. Recurrent methods take longer due to backpropagation through time, but this only happens during training. DWN has a superior performance to RCN in most cases, while also being faster to train. While PRN has better performance, when computational complexity is higher priority, our DWN can be used instead.

\subsubsection{Bee Waggle Dance data}
On this second real-world dataset, PRN outperforms other methods when $\eta\geq 5$ (around 1 second) with AUC of 93\% (fig. \ref{fig:auc-bee}). For tolerance of around 2 seconds ($\eta=2^6=64$), PRN has 93\% AUC while the next three best performing methods are RCN (84\%), CNN (79\%), and DWN (78\%) (see Appendix). 
Further, changes in tolerance affect our method much less than others. For instance, when the tolerance is lowered from 32 ($\eta=2^5=32$) to 16 ($\eta=2^4=16$), the AUC of RCN drops significantly (from 84\% to 18\%), while AUC of PRN drops much less (from 93\% to 61\%). CNN has a dramatic drop in accuracy from $\eta=16$ to $\eta=8$, suggesting it is consistently detecting changes with a delay. 
Thus, PRN is less sensitive to this parameter and more reliable for real world cases. 
Similar to the OPPORTUNITY data, GGM has the lowest AUC for all tolerances, and HM-RNN has the lowest AUC among deep learning methods.  Both PRN-S and HM-RNN have higher max AUC in Bee data (64\% and 57\%) than OPPORTUNITY (51\% and 30\%)  because the bee activities have more distinct boundaries than human ones. Thus, our PRN is more widely applicable.

\section{Conclusion}
We propose PRN, a new scale-invariant deep learning architecture, and show how it can detect from abrupt to gradual changepoints in multimodality time series. The core is 1) DWN: a CNN with trainable Wavelet layers that recognize short-term multi-scale patterns; and 2) PRL: a pyramid-shaped RNN on top of the multi-scale feature maps to simultaneously model long-term patterns and fuse multi-scale information. Unlike existing DNNs, PRN can detect events involving short- and long-term patterns and extrapolate to scales not seen in training. Experiments on real and synthetic data demonstrate that PRN detects changes quickly, with lower sensitivity to parameters than other approaches. Future work will focus on handling missing and noisy labels with semi-supervised learning methods.

\appendix

\section{Implementation details}
All of the deep-learning baselines share a core CNN architecture on which the additional modules are built. We fix the architecture of the core CNN to be $[9:128:4], [5:128:2], [5:128:2]$, where we use the notation $x:y:z$ for a convolution layer where $x$ is the kernel size, $y$ is the number of output feature maps, and $z$ is the pooling stride. Each convolution layer is followed by max-pooling and ReLU activation. The output of all baselines are fed to a fully connected perceptron with sigmoid activation which results in binary detection scores at each time step. The granularity ratio $\gamma$ for this architecture is $1/16$. 
For DWN, PRN, and PRN-S, we used a 7-level wavelet with kernel size 3 for both synthetic and OPPORTUNITY dataset. For Bee Waggle Dance data, due to the small size and more abrupt transitions, we used a 5-level wavelet with kernel size 3. For all datasets RCN and PRN used an LSTM cell with 256 hidden units, 128 units for HM-RNN. 

We train all models using Adam \citep{kingma2014adam} with early stopping to avoid overfitting with initial learning rate of 0.001. At test time, the models take a time series and predict a sequence of detection scores. To detect changepoints, we apply non-maximum suppression with a sliding window of length $\omega$ and filter the maximum values with a threshold. We evaluate AUC by iterating over this threshold. 
Hyper-parameter $\omega$ controls how nearby two distinct changes can be detected and is tuned for each method separately using cross-validation. 

The real world datasets (Bee data and OPPORTUNITY data) are more challenging than the synthetic data, as they include diverse changepoints formed by transitions between many activity types. To address this, we use multitask learning, training the model to both detect changes and classify activity by changing the output dimension of the last fully connected later to have multiple units (19 for OPPORTUNITY data, and 4 for Bee data). For OPPORTUNITY data, the first 18 units predict a log probability for each activity and the last 1 unit outputs the probability of a change point (for bee data, it's 3 units and 1 unit). We define a softmax cross-entropy loss on those 18 units and add it as a regularization term to the objective function. Multitask learning improved the results equally for all baselines, because the model has auxiliary information, namely the activity type and not just the existence of a change. 

For GGM, we use the full covariance model instead of the independent features model to capture the correlations between features. We use a uniform prior as in \citep{xuan2007modeling}, and set the pruning threshold to $10^{-20}$. Since there is no training for GGM, we evaluate the algorithm using the same test data as all other methods we compared on both synthetic and real world dataset.

We evaluate precision and recall, and report AUC. As detected changepoints may not exactly match the true changepoints, we use a tolerance parameter $\eta$ that sets how close a detected change must be to a true change to be considered a correct detection. We match detected changepoints to the closest true changepoint within $\eta$ time steps. 
Precision is the number of matched detections divided by the number of all detections, and recall is the number of matches divided by the number of true changes. 

\section{Results detail}

\begin{figure}
\centering 
\includegraphics[width=0.75\textwidth]{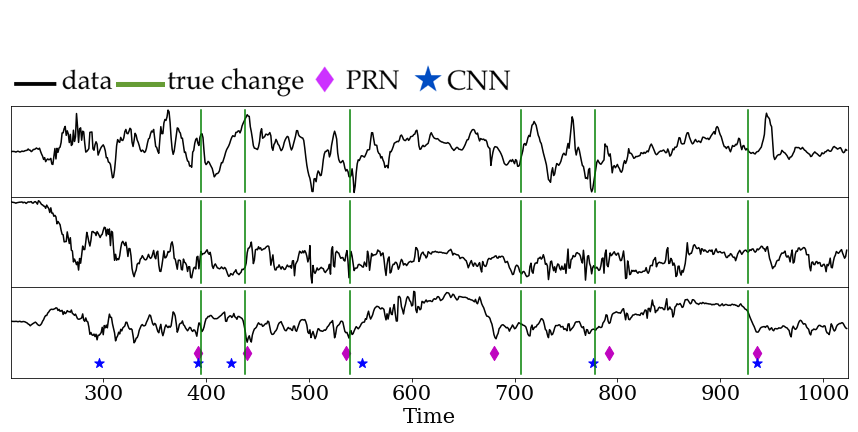}
\caption{\label{fig:opp} Detected changes on sample (3 of 79 dimensions shown) of OPPORTUNITY dataset.}
\end{figure}

\subsection{Synthetic data}

Tables \ref{tbl:syn-gradual}-\ref{tbl:syn-all} show the AUC (Area Under the ROC Curve) results for synthetic data. Table \ref{tbl:syn-gradual} shows the results for the experiment of ``train abrupt and test gradual", 
Table \ref{tbl:syn-abrupt} shows the results for the experiment of ``train gradual and test abrupt",
 and Table \ref{tbl:syn-all} shows the results for the experiment of ``train all and test all" for synthetic data.

\begin{table}[h!]
\caption{AUC results for the experiment of ``train abrupt and test gradual" for synthetic data for each tolerance ($\eta$)}
\label{tbl:syn-gradual}
\center
\begin{tabular}{l|lllllll}
 $\eta$& HM-RNN & RCN & CNN &GGM& PRN-S & DWN & PRN \\
 \hline
 8&\textbf{0.01}&0.002&0.003&0.007&0.006&0.005&0.003\\
16&0.01&0.008&0.011&0.010&\textbf{0.023}&0.016&0.011\\
32&0.01&0.026&0.042&0.012&\textbf{0.076}&0.063&0.043\\
64&0.011&0.107&0.153&0.012&0.271&\textbf{0.213}&0.155\\
128&0.013&0.391&0.421&0.013&\textbf{0.625}&0.546&0.503\\
256&0.064&0.783&0.757&0.016&0.849&0.822&\textbf{0.863}\\
512&0.120&0.824&0.801&0.023&0.862&0.835&\textbf{0.876}

\end{tabular}
\end{table}

\begin{table}[h!]
\caption{AUC results for the experiment of ``train gradual and test abrupt" for synthetic data for each tolerance ($\eta$)}
\label{tbl:syn-abrupt}
\center
\begin{tabular}{l|lllllll}
 $\eta$& HM-RNN & RCN & CNN &GGM& PRN-S & DWN & PRN \\
 \hline
 8&0.002&0.001&0.013&0.007&0.033&\textbf{0.102}&0.027\\
16&0.007&0.003&0.049&0.011&0.115&\textbf{0.301}&0.100\\
32&0.017&0.014&0.188&0.013&0.347&\textbf{0.599}&0.376\\
64&0.040&0.269&0.665&0.013&0.689&\textbf{0.787}&0.724\\
128&0.140&0.650&0.797&0.014&0.811&\textbf{0.822}&0.816\\
256&0.144&0.795&0.814&0.018&\textbf{0.833}&\textbf{0.833}&0.830\\
512&0.148&0.817&0.830&0.026&\textbf{0.834}&\textbf{0.834}&0.830

\end{tabular}
\end{table}

\begin{table}[h!]
\caption{AUC results for the experiment of ``train all and test all" for synthetic data for each tolerance ($\eta$)}
\label{tbl:syn-all}
\center
\begin{tabular}{l|lllllll}
 $\eta$& HM-RNN & RCN & CNN &GGM& PRN-S & DWN & PRN \\
 \hline
8&0.027&0.039&\textbf{0.061}&0.007&0.027&0.014&0.039\\
16&0.032&0.093&\textbf{0.144}&0.011&0.100&0.061&0.122\\
32&0.051&0.204&0.244&0.012&0.249&0.164&\textbf{0.284}\\
64&0.055&0.390&0.407&0.017&0.496&0.406&\textbf{0.551}\\
128&0.119&0.582&0.586&0.021&0.737&0.700&\textbf{0.747}\\
256&0.134&0.788&0.747&0.030&\textbf{0.863}&0.840&0.860\\
512&0.142&0.852&0.808&0.043&\textbf{0.874}&0.860&0.869

\end{tabular}
\end{table}

\subsection{Real World data}
\subsubsection{Opportunity data}
Table \ref{tbl:opp-results} shows the results for OPPORTUNITY data.
Figure \ref{fig:opp} shows example results for the OPPORTUNITY dataset tested using our scale invariant PRN and scale sensitive CNN. In the time series, we see that CNN has a missed detection and at least one false positive around time 300, while PRN detects all changes close to their actual times. Overall CNN has a higher false positive rate. While PRN's detected changes and ground truth are not always precisely aligned, the small gaps are acceptable in real world data, where it can be hard to define a single moment when the change occurs.

\begin{table}[h!]
\caption{AUC results of OPPORTUNITY data for each tolerance ($\eta$, with unit of 1/30 sec)}
\label{tbl:opp-results}
\center
\begin{tabular}{l|lllllll}
  $\eta$& HM-RNN & RCN & CNN &GGM& PRN-S & DWN & PRN \\
 \hline
 2&0.013&\textbf{0.036}&0.024&0.007&0.007&0.034&0.040\\
4&0.028&0.077&0.066&0.016&0.024&0.093&\textbf{0.104}\\
8&0.053&0.213&0.197&0.022&0.068&0.234&\textbf{0.297}\\
16&0.143&0.513&0.442&0.027&0.236&0.515&\textbf{0.601}\\
32&0.187&0.713&0.629&0.032&0.418&0.712&\textbf{0.773}\\
64&0.240&0.744&0.687&0.046&0.471&0.753&\textbf{0.815}\\
128&0.300&0.771&0.710&0.065&0.507&0.759&\textbf{0.833}

\end{tabular}
\end{table}

\subsubsection{Bee Waggle Dance data}
Table \ref{tbl:bee-results} shows the results for Bee Waggle Dance data. The results show that our proposed PRN has the highest AUC score for almost all tolerance values (except $\eta=2$). Thus, our proposed PRN is less sensitive to the tolerance and more reliable for real world cases.

\begin{table}[h!]
\caption{AUC results of Bee Waggle Dance data for each tolerance ($\eta$, with unit of 1/30 sec)}
\label{tbl:bee-results}
\center
\begin{tabular}{l|lllllll}
 $\eta$& HM-RNN & RCN & CNN &GGM& PRN-S & DWN & PRN \\
 \hline
2&\textbf{0.033}&0.007&0.008&0.019&0.009&0.007&0.025\\
4&0.144&0.007&0.053&0.023&0.009&0.023&\textbf{0.145}\\
8&0.362&0.054&0.192&0.041&0.119&0.131&\textbf{0.400}\\
16&0.480&0.178&0.703&0.077&0.393&0.329&\textbf{0.608}\\
32&0.573&0.841&0.789&0.083&0.643&0.777&\textbf{0.932}\\
64&0.573&0.841&0.789&0.083&0.643&0.777&\textbf{0.932}\\
128&0.573&0.841&0.789&0.083&0.643&0.777&\textbf{0.932}

\end{tabular}
\end{table}

\bibliographystyle{named}
\bibliography{my_bib}

\end{document}